\documentclass[a4paper,fleqn]{cas-dc}

\usepackage[numbers]{natbib}
\usepackage{amsmath}
\usepackage{amsfonts}
\usepackage{amssymb}
\usepackage{algorithm}
\usepackage{algpseudocode}
\usepackage{graphicx}
\usepackage{colortbl}
\usepackage[table]{xcolor} 

\usepackage{subcaption}
\usepackage{float}
\usepackage{stfloats}
\usepackage{booktabs}
\usepackage{multirow}

\definecolor{bestblue}{RGB}{198, 226, 255}   
\definecolor{secondgray}{RGB}{224, 224, 224} 

\usepackage{ragged2e}

\def\tsc#1{\csdef{#1}{\textsc{\lowercase{#1}}\xspace}}
\tsc{WGM}
\tsc{QE}
\tsc{EP}
\tsc{PMS}
\tsc{BEC}
\tsc{DE}

\begin{document}
\let\WriteBookmarks\relax
\def\floatpagepagefraction{1}
\def\textpagefraction{.001}

\shorttitle{Reasoning under Ambiguity: Uncertainty-Aware Multilingual Emotion Classification under Partial Supervision}

\shortauthors{M. M. Hossain et~al.}

\title [mode = title]{Reasoning under Ambiguity: Uncertainty-Aware Multilingual Emotion Classification under Partial Supervision}

\author[1]{Md. Mithun Hossain}[orcid=0009-0001-4883-1802]\cormark[1]

\ead{mithunhossain@bubt.edu.bd}

\credit{Conceptualization of this study, Methodology, Data Curation, Software, Writing - Original Draft}

\author[2]{Mashary N. Alrasheedy}[orcid=0000-0002-4253-1710]
\ead{mn.alrasheedy@uoh.edu.sa}

\credit{Investigation, Methodology, Formal analysis, Visualization, Writing - Original Draft}

\author[3]{Nirban Bhowmick}[orcid=0009-0001-9561-486X]
\ead{nirban@scholarx.academy}

\credit{Investigation, Formal analysis, Visualization, Writing - Original Draft}

\author[4]{Shamim Forhad}[orcid=0000-0001-7449-6182]
\ead{shamim@uttara.ac.bd}

\credit{Validation, Supervision, Writing - Original Draft, Writing - Review \& Editing}

\author[5]{Md. Shakil Hossain}[orcid=0009-0009-1584-3282]
\ead{shakilhossain@bubt.edu.bd}

\credit{Validation, Supervision, Writing - Original Draft, Writing - Review \& Editing}

\author[5]{Sudipto Chaki}[orcid=0000-0002-7286-6722]
\ead{sudipto@bubt.edu.bd}

\credit{Resources, Funding acquisition, Writing - Review \& Editing}

\author[5]{Md Shafiqul Islam}[orcid=0000-0001-9469-2041]
\ead{msislam@bubt.edu.bd}

\credit{Resources, Funding acquisition, Writing - Review \& Editing}

\affiliation[1]{organization={BUBT Research Graduate School, Bangladesh University of Business and Technology},
            city={Dhaka},
            postcode={1216}, 
            country={Bangladesh}}

\affiliation[2]{organization={Department of Computer Science Applied College, University of Ha’il},
            city={Ha’il},
            country={Saudi Arabia}}
\affiliation[3]{organization={Electrical and Computer Engineering, University of Central Florida},
            addressline={Orlando}, 
            city={FL},
            postcode={32816}, 
            country={USA}}

\affiliation[4]{organization={Electrical and Computer Engineering, Uttara University},
            addressline={77 Beribadh Road}, 
            city={Dhaka},
            postcode={1230}, 
            country={Bangladesh}}
\affiliation[5]{organization={Department of Computer Science and Engineering, Bangladesh University of Business and Technology},
            city={Dhaka},
            postcode={1216}, 
            country={Bangladesh}}

\cortext[1]{Corresponding author: M. M. Hossain (mithunhossain@bubt.edu.bd)}

\begin{abstract}
Contemporary knowledge-based systems increasingly rely on multilingual emotion identification to support intelligent decision-making, yet they face major challenges due to emotional ambiguity and incomplete supervision. Emotion recognition from text is inherently uncertain because multiple emotional states often co-occur and emotion annotations are frequently missing or heterogeneous. Most existing multi-label emotion classification methods assume fully observed labels and rely on deterministic learning objectives, which can lead to biased learning and unreliable predictions under partial supervision. This paper introduces Reasoning under Ambiguity, an uncertainty-aware framework for multilingual multi-label emotion classification that explicitly aligns learning with annotation uncertainty. The proposed approach uses a shared multilingual encoder with language-specific optimization and an entropy-based ambiguity weighting mechanism that down-weights highly ambiguous training instances rather than treating missing labels as negative evidence. A mask-aware objective with positive-unlabeled regularization is further incorporated to enable robust learning under partial supervision. Experiments on English, Spanish, and Arabic emotion classification benchmarks demonstrate consistent improvements over strong baselines across multiple evaluation metrics, along with improved training stability, robustness to annotation sparsity, and enhanced interpretability.
\end{abstract}


\maketitle

\section{Introduction}
A key component of many knowledge-based systems, including social media monitoring, opinion mining, mental health evaluation, and human-computer interaction, is the classification of emotions from textual input. Emotion recognition is intrinsically multi-label \cite{hossain2025emonet}, in contrast to traditional sentiment analysis, which concentrates on a single polarity dimension \cite{hossain2025enhancing, hossain2025dynamic}, since a single utterance may concurrently express multiple emotional states. Early work on affective text analysis and large-scale benchmarks such as SemEval-2018 Affect in Tweets revealed that emotion co-occurrence and severe class imbalance are intrinsic properties of real-world emotion data \cite{mohammad2018semeval,mohammad2017emotion}. These characteristics make emotion classification particularly challenging, as models must reason over overlapping labels while capturing subtle affective cues embedded in natural language.

Performance in emotion categorization has significantly increased due to recent developments in pretrained language models. Effective contextual representation learning across languages has been made possible by models like BERT \cite{devlin2019bert}, RoBERTa \cite{liu2019roberta}, and their multilingual equivalents, such as multilingual BERT and XLM-R \cite{conneau2020xlmr}. These models are commonly adapted to multi-label emotion classification by applying independent sigmoid classifiers for each emotion category \cite{bostan2018analysis,demszky2020goemotions}. Nevertheless, the majority of current methods require fully observable emotion labels during training and depend on deterministic prediction algorithms. This presumption restricts their application in knowledge-driven decision-support systems because it ignores the inherent ambiguity of emotional language and frequently results in overconfident forecasts that are poorly calibrated and challenging to understand.
The limitations of deterministic multi-label learning become more pronounced in multilingual settings. Emotion datasets are typically constructed independently for each language and frequently provide heterogeneous label inventories, resulting in partially annotated data when mapped into a unified label space. In such cases, missing labels do not necessarily indicate negative evidence but rather reflect annotation incompleteness. Prior studies on weakly supervised and partially labeled learning have shown that ignoring missing labels or naively treating them as negatives can introduce significant bias \cite{zhou2018weak}. In multilingual emotion classification, this issue is further exacerbated by linguistic and cultural variation in emotional expression \cite{fatima2024language}, which increases ambiguity at both the instance and label levels. As a result, models trained under partial supervision may exhibit degraded generalization and unreliable confidence estimates.

Motivated by these challenges, this work addresses the problem of reasoning under ambiguity in multi-label emotion classification with partial supervision. We propose an uncertainty-aware learning framework that explicitly models prediction ambiguity and incomplete annotations while emphasizing language-specific training to preserve language-dependent emotional characteristics. A shared multilingual encoder is employed solely as a representation backbone, while uncertainty is incorporated at the learning and inference stages. Multilingual joint training is examined only as an ablation study to analyze cross-lingual effects. Through comprehensive experiments on English and Chinese emotion datasets, we demonstrate that explicitly modeling uncertainty leads to more robust, better-calibrated, and more interpretable emotion predictions, which are essential properties for reliable knowledge-based emotion analysis systems.

\paragraph{Contributions:}
The main contributions of this work are summarized as follows:
\begin{itemize}
    \item We propose an uncertainty-aware multi-label emotion classification framework that explicitly models linguistic ambiguity and partial supervision, addressing key limitations of deterministic emotion classifiers.
    \item We introduce an ambiguity-aware reweighting strategy based on entropy measures that dynamically down-weights highly uncertain training instances, improving robustness to ambiguous emotional expressions.
    \item We develop an evidential learning formulation for multi-label emotion classification that captures label-wise uncertainty through Beta distributions, enabling principled uncertainty estimation.
    \item We design a mask-aware learning objective augmented with positive-unlabeled regularization, allowing effective learning from missing emotion annotations without introducing spurious negative supervision.
\end{itemize}
The remainder of this paper is organized as follows. Section~\ref{related_work} reviews related work on emotion classification, multi-label learning, and uncertainty modeling. Section~\ref{sec:method} introduces the proposed ambiguity-aware learning framework. Section~\ref{Experiment_results} presents the experimental setup and a detailed analysis of results across multiple languages. Section~\ref{discussion} provides a broader discussion of the findings and their implications. Section~\ref{Limitations_futurework} discusses the limitations of the proposed approach and outlines directions for future research, while Section~\ref{conclusion} concludes the paper.

\section{Related Works}\label{related_work}
\paragraph{Emotion Classification and Datasets:}
The SemEval-2018 Task 1 benchmark and associated affect datasets offer standardized assessment frameworks for English, Spanish, and Arabic tweets \cite{mohammad2018semeval,mohammad2017emotion}. Early research on emotion analysis in text relied on discrete emotion classification and intensity prediction. The emotion inventory was extended by later datasets, such as GoEmotions, which emphasized the frequency of co-occurring emotions in brief texts \cite{demszky2020goemotions}. Prior analyses of emotion corpora have also emphasized annotation subjectivity, label overlap, and disagreement as inherent properties of emotion data rather than annotation noise \cite{bostan2018analysis,poria2017review}. These observations motivate learning settings that go beyond single-label assumptions and account for partial and ambiguous supervision.

\paragraph{Multi-Label Learning Methods:}
With traditional methods like ML-KNN, classifier chains, and label correlation modeling establishing fundamental concepts, multi-label classification has been thoroughly researched \cite{zhang2007ml,read2009classifier,read2011classifier,zhang2014review}. More contemporary neural techniques, such as Seq2Set, SGM, and set prediction networks \cite{yang2018sgm,zhang2019dspn}, reformulate multi-label prediction as sequence or set generation problems. Further enabling permutation-invariant label modeling were attention-based and transformer-style architectures \cite{lee2019set,vaswani2017attention}. However, most of these methods assume fully observed labels during training and implicitly treat missing labels as negative evidence, which can be problematic for emotion datasets with incomplete annotations. 

\paragraph{Neural Models for Emotion Classification:}
Deep learning models for emotion classification have progressed from CNNs and recurrent networks to massive pretrained transformers \cite{kim2014cnn,schuster1997bilstm,devlin2019bert,liu2019roberta,conneau2020xlmr}. Several studies \cite{zhou2016attention,zhang2021spanemo} have used span-based formulations or attention approaches to better capture emotionally important text portions. Graph-based techniques use GCNs or GATs to model label or word connections (Velickovic 2018; Yao 2019; Chochlakis 2022). Although these models boost representational capacity, they typically optimize simple multi-label losses without explicitly addressing annotation ambiguity.

\paragraph{Uncertainty, Weak Supervision, and Ambiguous Labels:}
In situations of weak supervision and ambiguous labeling, typified by several candidate labels or inadequate supervision, the investigation of learning under uncertainty has been studied \cite{zhou2018weak}. Evidential and distribution-based learning techniques in emotion categorization attempt to capture prediction-level uncertainty (Wu, 2023). Lin et al. \cite{lin2017focal,lin2022contrastive} studied focal loss and contrastive learning to minimize imbalance and noise in supervision. However, rather than uncertainty caused by missing annotations, these methods often focus on output ambiguity or class imbalance. As shown in recent analyses, predictive uncertainty alone may be insufficient when label absence does not imply label negation.

\paragraph{Positioning of Our Work:}
In contrast to prior approaches, our work explicitly models annotation ambiguity by leveraging entropy derived from observed label distributions and integrating it into the training objective via instance-level weighting. Rather than modifying the prediction space or assuming complete supervision, we align uncertainty modeling with the annotation process itself. This design allows the model to down-weight highly ambiguous instances while preserving informative but partially labeled data. Our framework is complementary to existing neural architectures and multi-label learning methods and can be applied without architectural changes. By demonstrating consistent gains across languages, improved training stability, and interpretable embedding behavior, our work bridges multi-label learning and ambiguity-aware supervision for multilingual emotion classification.

\section{Proposed Methodology}
\label{sec:method}

\subsection{Problem Formulation}

We study the task of multilingual multi-label emotion classification under partial supervision.
Given an input text $x_i$ written in language $\ell \in \{\text{EN}, \text{ES}, \text{AR}\}$, the objective is to predict a set of emotion labels from a unified label space of size $K$.
Each training instance is associated with a binary label vector $\mathbf{y}_i \in \{0,1\}^K$ and a label mask $\mathbf{m}_i \in \{0,1\}^K$, where $m_{ik}=1$ indicates that label $k$ is observed and $m_{ik}=0$ denotes a missing annotation.
This formulation captures realistic annotation scenarios in which only a subset of emotions is explicitly labeled per instance.

\subsection{Multilingual Representation Learning}

\begin{figure*}[t]
    \centering
    \includegraphics[width=\linewidth]{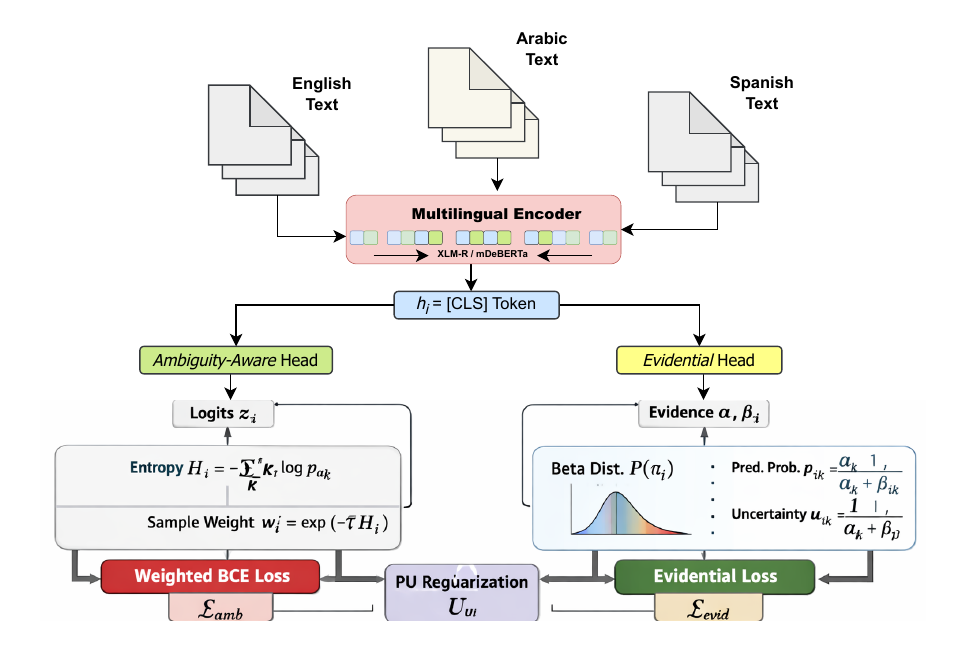}
    \caption{\small 
    Overview of the proposed \textbf{Reasoning under Ambiguity} framework.
    Multilingual inputs (English, Spanish, Arabic) are encoded using a shared multilingual encoder (XLM-R or mDeBERTa).
    The \texttt{[CLS]} representation is fed into an ambiguity-aware head that estimates prediction entropy and assigns
    sample-level weights for ambiguity-weighted learning under partial supervision.
    An evidential head is included for comparative uncertainty modeling.
    Missing labels are handled using a mask-aware objective with positive--unlabeled regularization.
    }
    \label{fig:architecture}
\end{figure*}

As illustrated in \textcolor{blue}{Figure~\ref{fig:architecture}}, we employ a shared multilingual encoder (XLM-R or mDeBERTa) to learn contextual representations across languages.
Given an input text $x_i$, a language identifier token is prepended before encoding, allowing the model to distinguish language-specific emotional expressions while sharing parameters across languages.
The encoder produces a sequence of contextualized representations:
\begin{equation}
\mathbf{H}_i = \text{Encoder}(x_i),
\label{eq:encoder}
\end{equation}
where $\mathbf{H}_i = \{\mathbf{h}_{i1}, \mathbf{h}_{i2}, \ldots, \mathbf{h}_{iT}\}$ and $T$ denotes the sequence length.
We use the contextual representation of the special \texttt{[CLS]} token as a sentence-level embedding:
\begin{equation}
\mathbf{h}_i = \mathbf{H}_{i,\texttt{[CLS]}} \in \mathbb{R}^{d},
\label{eq:cls}
\end{equation}
where $d$ is the hidden dimensionality of the encoder.
This embedding serves as a compact representation capturing both semantic meaning and emotion-related cues, and forms the basis for subsequent prediction.

\subsection{Ambiguity-Aware Multi-Label Prediction}

Emotion expressions often exhibit inherent ambiguity, with multiple emotions co-occurring or overlapping in a single utterance.
To explicitly account for this phenomenon, we introduce an ambiguity-aware prediction mechanism that estimates uncertainty at the instance level.
\subsubsection{Ambiguity-Aware Prediction Head}

The sentence-level representation $\mathbf{h}_i$, obtained from the multilingual encoder, is mapped to emotion-specific prediction scores through a linear multi-label prediction head.
Formally, the logits are computed as:
\begin{equation}
\mathbf{z}_i = \mathbf{W}\mathbf{h}_i + \mathbf{b},
\label{eq:logits}
\end{equation}
where $\mathbf{W} \in \mathbb{R}^{K \times d}$ is a weight matrix that projects the $d$-dimensional sentence embedding into the $K$-dimensional emotion space, and $\mathbf{b} \in \mathbb{R}^{K}$ is a bias term.
Each logit $z_{ik}$ represents the unnormalized confidence of emotion label $k$ being expressed in the input text $x_i$.
To obtain probabilistic predictions, we apply a sigmoid activation function independently to each logit:
\begin{equation}
\mathbf{p}_i = \sigma(\mathbf{z}_i),
\label{eq:sigmoid}
\end{equation}
where $p_{ik} \in [0,1]$ denotes the predicted probability that emotion $k$ is present in the input.
Unlike a softmax function, which enforces mutual exclusivity among classes, the sigmoid activation allows multiple emotions to be activated simultaneously.
This independent-label formulation is particularly well-suited for emotion classification, as emotional states frequently co-occur (e.g., joy and surprise, or anger and sadness) and do not obey a single-label assumption.
Moreover, treating each label independently enables the model to capture fine-grained emotion patterns without imposing artificial competition between labels.

Importantly, the probabilistic outputs $\mathbf{p}_i$ in Eq.~\ref{eq:sigmoid} serve a dual role in our framework.
In addition to supporting standard multi-label prediction, they form the basis for subsequent ambiguity estimation, where prediction entropy is computed over observed labels to quantify instance-level uncertainty.
Thus, the linear prediction head not only produces emotion scores but also facilitates principled uncertainty-aware learning under partial supervision.

\subsubsection{Entropy-Based Ambiguity Estimation}

To quantify prediction ambiguity, we compute the entropy of the predicted probabilities, restricted to observed labels only.
Let $\mathcal{O}_i = \{k \mid m_{ik} = 1\}$ denote the set of observed labels for instance $i$.
The normalized entropy is defined as:
\begin{equation}
H_i = -\frac{1}{|\mathcal{O}_i|}
\sum_{k \in \mathcal{O}_i}
\Big[
p_{ik} \log p_{ik}
+
(1 - p_{ik}) \log (1 - p_{ik})
\Big].
\label{eq:entropy}
\end{equation}
As shown in Eq.~\ref{eq:entropy}, higher entropy values correspond to greater uncertainty in the model’s predictions, reflecting ambiguous emotional content or conflicting cues in the input text.

\subsubsection{Ambiguity-Weighted Learning Objective}

We transform the entropy score in Eq.~\ref{eq:entropy} into a sample-level ambiguity weight:
\begin{equation}
w_i = \exp(-\tau H_i),
\label{eq:weight}
\end{equation}
where $\tau$ is a temperature hyperparameter controlling the sensitivity of the weighting mechanism.
Lower weights are assigned to highly ambiguous instances, preventing them from disproportionately influencing the optimization process.
Using the ambiguity weight in Eq.~\ref{eq:weight}, we define the ambiguity-weighted masked binary cross-entropy loss:
\begin{equation}
\mathcal{L}_{\text{amb}} =
\frac{1}{N}
\sum_{i=1}^{N}
w_i
\cdot
\frac{
\sum_{k=1}^{K}
m_{ik}\,\text{BCE}(z_{ik}, y_{ik})
}{
\sum_{k=1}^{K} m_{ik}
}.
\label{eq:amb_loss}
\end{equation}

Eq.~\ref{eq:amb_loss} ensures that learning is driven primarily by confident and informative samples while remaining robust to ambiguity and annotation noise.

\subsection{Learning with Missing Labels}
In partially supervised settings, treating missing labels as true negatives can introduce systematic bias.
To mitigate this issue, we incorporate a positive--unlabeled (PU) regularization term that softly constrains predictions for unobserved labels.
For labels where $m_{ik}=0$, we apply a weak negative penalty:
\begin{equation}
\mathcal{L}_{\text{PU}} =
\frac{1}{N}
\sum_{i=1}^{N}
\sum_{k=1}^{K}
(1 - m_{ik})\,\text{BCE}(z_{ik}, 0).
\label{eq:pu_loss}
\end{equation}

As shown in Eq.~\ref{eq:pu_loss}, this regularization discourages overly confident positive predictions for missing labels without enforcing incorrect negative supervision.

\subsection{Overall Training Objective}

The final training objective combines ambiguity-weighted supervision and PU regularization:
\begin{equation}
\mathcal{L} =
\mathcal{L}_{\text{amb}}
+
\lambda_{\text{PU}}\,\mathcal{L}_{\text{PU}},
\label{eq:total_loss}
\end{equation}
where $\lambda_{\text{PU}}$ controls the contribution of the PU regularization term. Eq.~\ref{eq:total_loss} defines the objective optimized during training.

\subsection{Inference}

At inference time, the model outputs emotion probabilities $\mathbf{p}_i$ as defined in Eq.~\ref{eq:sigmoid}.
Label-specific decision thresholds are tuned on the validation set and applied to obtain final multi-label predictions.

\begin{algorithm*}[t]
\caption{Reasoning under Ambiguity: Ambiguity-Weighted Learning with Partial Supervision}
\label{alg:ambiguity}
\footnotesize
\small
\begin{algorithmic}[1]
\Require
Training set $\mathcal{D}=\{(x_i, \mathbf{y}_i, \mathbf{m}_i)\}_{i=1}^{N}$,
multilingual encoder $E(\cdot)$,
number of labels $K$,
temperature $\tau$,
PU weight $\lambda_{\text{PU}}$

\Ensure
Trained model parameters $\theta$

\State Initialize encoder and ambiguity-aware head parameters $\theta$

\For{each training epoch}
    \For{each mini-batch $\mathcal{B} \subset \mathcal{D}$}
        \State Encode texts: $ \mathbf{h}_i \leftarrow E(x_i), \quad \forall (x_i,\mathbf{y}_i,\mathbf{m}_i) \in \mathcal{B} $
        \State Compute logits and probabilities: 
       $ \mathbf{z}_i \leftarrow \mathbf{W}\mathbf{h}_i + \mathbf{b},
        \quad
        \mathbf{p}_i \leftarrow \sigma(\mathbf{z}_i)
        $
        \State Compute entropy over observed labels:
        $
        H_i \leftarrow
        -\frac{1}{|\mathcal{O}_i|}
        \sum_{k \in \mathcal{O}_i}
        \big(
        p_{ik}\log p_{ik} + (1-p_{ik})\log(1-p_{ik})
        \big)
        $
        \State Compute ambiguity weights:
        $
        w_i \leftarrow \exp(-\tau H_i)
        $
        \State Compute ambiguity-weighted masked BCE loss:
        $
        \mathcal{L}_{\text{amb}} \leftarrow
        \frac{1}{|\mathcal{B}|}
        \sum_{i \in \mathcal{B}}
        w_i
        \cdot
        \frac{
        \sum_{k=1}^{K} m_{ik}\,
        \text{BCE}(z_{ik}, y_{ik})
        }{
        \sum_{k=1}^{K} m_{ik}
        }
        $
        \State Compute PU regularization loss:
        $
        \mathcal{L}_{\text{PU}} \leftarrow
        \frac{1}{|\mathcal{B}|}
        \sum_{i \in \mathcal{B}}
        \sum_{k=1}^{K}
        (1-m_{ik})\,\text{BCE}(z_{ik},0)
        $
        \State Compute total loss:
        $
        \mathcal{L} \leftarrow
        \mathcal{L}_{\text{amb}} + \lambda_{\text{PU}}\mathcal{L}_{\text{PU}}
        $
        \State Update parameters $\theta$ using gradient descent
    \EndFor
\EndFor

\State \Return trained parameters $\theta$
\end{algorithmic}
\end{algorithm*}

\subsection{Ambiguity-Weighted Learning Training Algorithm}
\textcolor{blue}{Algorithm~\ref{alg:ambiguity}} outlines the proposed ambiguity-aware learning procedure under partial supervision. Each training instance is associated with a binary label vector and a corresponding mask indicating which emotion labels are observed. During training, a multilingual encoder produces instance representations that are mapped to label probabilities, from which an entropy-based ambiguity score is computed over observed labels only. This entropy controls an instance-level weight that down-weights highly ambiguous examples. The main objective is a masked binary cross-entropy loss scaled by the ambiguity weight, ensuring that learning focuses on reliable supervision while preserving partially labeled data. An optional positive–unlabeled regularization term further constrains predictions on unobserved labels without treating them as negative evidence. Together, these components enable stable and effective optimization in the presence of incomplete and ambiguous emotion annotations.

\section{Experiment and Result Analysis}\label{Experiment_results}
All experiments were conducted in a Python-based deep learning environment built on PyTorch 2.9 with CUDA 12.6 support, running on a single NVIDIA RTX 2060 GPU with 12 GB of VRAM. Model development and data processing relied on the HuggingFace ecosystem, including \textit{transformers} (v4.57), \textit{datasets} (v4.4), \textit{accelerate} (v1.12), and \textit{huggingface-hub} (v0.36). Multilingual encoders such as XLM-R and mDeBERTa were implemented using sentence-transformers, while uncertainty-aware learning and evaluation employed standard scientific libraries including NumPy (v2.3), SciPy (v1.16), and scikit-learn (v1.8). Memory-efficient training was supported via mixed-precision computation and bitsandbytes, enabling stable multi-seed experiments within limited GPU resources. Visualization and analysis used Matplotlib and Seaborn, and nearest-neighbor interpretability analyses were conducted using cosine similarity with UMAP and PyNNDescent. All experiments were executed on a Linux-based system with properly configured NVIDIA CUDA libraries (cuDNN, cuBLAS, NCCL), ensuring reproducibility across runs.
\subsection{Datasets}

\begin{table*}[t]
\centering
\caption{\small Dataset statistics by language and split. All floating-point values are reported with two decimal places.}
\label{tab:dataset_stats}
\small
\setlength{\tabcolsep}{4pt}
\renewcommand{\arraystretch}{1.15}
\begin{tabular}{l l r r r r r r}
\toprule
\textbf{Lang} & \textbf{Split} & \textbf{\#Inst} & \textbf{AvgTok} & \textbf{MedTok} & \textbf{\#Labels} & \textbf{AvgLbl/Inst} & \textbf{\%Multi} \\
\midrule
EN & train      & 6838 & 16.06 & 17.00 & 11 & 2.35 & 82.66 \\
EN & validation &  886 & 15.86 & 16.00 & 11 & 2.44 & 85.21 \\
EN & test       & 3259 & 15.89 & 16.00 & 11 & 2.41 & 85.98 \\
\addlinespace[2pt]
ES & train      & 3559 & 13.54 & 13.00 & 11 & 1.67 & 56.28 \\
ES & validation &  679 & 13.07 & 12.00 & 11 & 1.65 & 54.64 \\
ES & test       & 2854 & 13.38 & 12.00 & 11 & 1.66 & 55.19 \\
\addlinespace[2pt]
AR & train      & 2278 & 16.51 & 17.00 & 11 & 2.27 & 77.96 \\
AR & validation &  585 & 16.26 & 16.00 & 11 & 2.37 & 80.51 \\
AR & test       & 1518 & 16.61 & 17.00 & 11 & 2.34 & 78.99 \\
\bottomrule
\end{tabular}
\end{table*}
We evaluate our framework on the SemEval-2018 Task 1 Emotion Classification (E-C) dataset \cite{mohammad2018semeval}, which provides multi-label emotion annotations for short, user-generated texts. We consider three languages, namely English (EN), Spanish (ES), and Arabic (AR), and additionally analyze their multilingual combination (BOTH). Each instance may express multiple co-occurring emotions drawn from a unified inventory of 11 emotion categories. Due to differences in annotation coverage across languages and splits, the dataset exhibits partial supervision, where only a subset of emotion labels is observed for each instance and unannotated labels do not necessarily indicate negative evidence. This setting reflects realistic multilingual emotion analysis scenarios with incomplete and heterogeneous annotations. To support principled learning under such conditions, we explicitly model label masks during training and avoid treating missing labels as negative supervision. The datasets are evaluated both individually and jointly to examine language-specific behavior and cross-lingual effects under ambiguity-aware learning.

\textcolor{blue}{Table~\ref{tab:dataset_stats}} summarizes dataset statistics by language and split. Texts in all languages are short, with similar average token lengths, albeit English and Arabic are significantly longer than Spanish.The same label inventory is used for all splits, however the average number of labels per instance changes by language, demonstrating different amounts of emotional co-occurrence. The same label inventory is used for all splits, however the average number of labels per instance changes by language, demonstrating different amounts of emotional co-occurrence.  These characteristics emphasize considerable label overlap and language diversity, promoting ambiguity-aware multi-label learning with minimal supervision.
\subsection{Implementation Details}
\label{subsec:impl}

We implement all models in \texttt{PyTorch} using the \texttt{Transformers} library.
For multilingual representation learning, we use a shared pretrained encoder (default: XLM-R) and fine-tune it for multi-label emotion classification with a sigmoid output layer (Eq.~\ref{eq:sigmoid}).
All experiments are conducted in a language-specific training setting (EN/ES/AR), and we report mean $\pm$ std over three random seeds for ablations.
For partial supervision, we apply a mask-aware objective that ignores missing labels during loss computation, and our main uncertainty mechanism is ambiguity-weighted learning, which down-weights highly ambiguous instances based on entropy-derived uncertainty (Sec.~\ref{sec:method}).

We optimize using AdamW with linear learning-rate warmup followed by decay.
Unless stated otherwise, we select checkpoints based on the development-set micro-F1 and then evaluate once on the test set using the same decision thresholding strategy.
Table~\ref{tab:hyperparams} summarizes the hyperparameters used across languages.

\begin{table}[t]
\centering
\small
\caption{\small Hyperparameters used in our experiments (unless otherwise specified).}
\setlength{\tabcolsep}{5pt}
\renewcommand{\arraystretch}{1.15}
\resizebox{\columnwidth}{!}{
\begin{tabular}{l c}
\toprule
\textbf{Hyperparameter} & \textbf{Value} \\
\midrule
Encoder backbone & \texttt{xlm-roberta-base} (default) \\
Max sequence length & 192 \\
Dropout & 0.10 \\
Batch size & 16 \\
Optimizer & AdamW \\
Learning rate & $2 \times 10^{-5}$ \\
Weight decay & 0.01 \\
Warmup ratio & 0.06 \\
Epochs & 5 \\
Gradient clipping & 1.0 \\
Loss & masked BCE (multi-label) \\
Thresholding & fixed 0.5 or dev-tuned global threshold \\
Ambiguity weighting temperature $\tau$ & 2.0 \\
Random seeds & \{42, 123, 2025\} \\
\bottomrule
\end{tabular}}
\label{tab:hyperparams}
\end{table}

\begin{table*}[t]
\centering
\caption{\small Performance comparison across \textbf{English}, \textbf{Arabic}, and \textbf{Spanish} emotion classification datasets.
Lower is better for HL and RL; higher is better for miF1, maF1, AP, and jacS.}
\label{tab:multilingual_results}
\small
\setlength{\tabcolsep}{4pt}
\renewcommand{\arraystretch}{1.12}
\begin{tabular}{l ccccc ccc ccc}
\toprule
\multirow{2}{*}{\textbf{Model}} 
& \multicolumn{5}{c}{\textbf{English}} 
& \multicolumn{3}{c}{\textbf{Arabic}} 
& \multicolumn{3}{c}{\textbf{Spanish}} \\
\cmidrule(lr){2-6} \cmidrule(lr){7-9} \cmidrule(lr){10-12}
& HL $\downarrow$ & RL $\downarrow$ & miF1 $\uparrow$ & maF1 $\uparrow$ & AP $\uparrow$
& miF1 $\uparrow$ & maF1 $\uparrow$ & jacS $\uparrow$
& miF1 $\uparrow$ & maF1 $\uparrow$ & jacS $\uparrow$ \\
\midrule
ECC \cite{read2009classifier}        & 0.210 & 0.240 & 0.458 & 0.376 & 0.395 & -- & -- & -- & -- & -- & -- \\
MLLOC \cite{huang2012multi}          & 0.245 & 0.342 & 0.484 & 0.414 & 0.413 & -- & -- & -- & -- & -- & -- \\
ML-KNN \cite{zhang2014review}        & 0.196 & 0.270 & 0.410 & 0.387 & 0.391 & -- & -- & -- & -- & -- & -- \\
TMC \cite{wang2016multi}             & 0.191 & 0.219 & 0.561 & 0.465 & 0.482 & -- & -- & -- & -- & -- & -- \\
SGM \cite{yang2018sgm}               & 0.165 & 0.184 & 0.616 & 0.492 & 0.524 & -- & -- & -- & -- & -- & -- \\
RERc \cite{zhou2018emotion}          & 0.176 & 0.170 & 0.651 & 0.539 & 0.530 & -- & -- & -- & -- & -- & -- \\
DATN \cite{yu2018improving}          & --    & --    & --    & 0.551 & --    & -- & -- & -- & -- & -- & -- \\
LEM \cite{fei2020latent}                                 & \cellcolor{secondgray}0.142 & \cellcolor{secondgray}0.157 & 0.675 & 0.567 & \cellcolor{secondgray}0.568 & -- & -- & -- & -- & -- & -- \\
SpanEmo \cite{alhuzali2021spanemo}                             & -- & -- & 0.713 & \cellcolor{secondgray}0.578 & -- 
                                     & 0.666 & 0.521 & \cellcolor{secondgray}0.548
                                     & \cellcolor{secondgray}0.641 & 0.532 & 0.532 \\
\midrule
Tw-StAR \cite{mulki2018twstar}                             & -- & -- & -- & -- & -- & 0.597 & 0.446 & 0.465 & 0.520 & 0.392 & 0.438 \\
EMA \cite{badaro2018ema}                              & -- & -- & -- & -- & -- & 0.618 & 0.461 & 0.489 & -- & -- & -- \\
ELiRF \cite{gonzalez2018elirf}                               & -- & -- & -- & -- & -- & -- & -- & -- & 0.535 & 0.440 & 0.458 \\
MILAB \cite{mohammad2018milab}                               & -- & -- & -- & -- & -- & -- & -- & -- & 0.558 & 0.407 & 0.469 \\
BERT\textsubscript{base} \cite{devlin2019bert}             & -- & -- & -- & -- & -- & 0.650 & 0.477 & 0.523 & 0.596 & 0.474 & 0.487 \\
HEF \cite{alswaidan2020hef}                                 & -- & -- & -- & -- & -- & 0.631 & 0.502 & 0.512 & -- & -- & -- \\
JSCL \cite{lin2022jscl}                                & -- & -- & -- & -- & -- & 0.660 & 0.540 & 0.541 & 0.634 & 0.556 & 0.529 \\
JSPCL \cite{lin2022contrastive}                                & -- & -- & -- & -- & -- & 0.658 & 0.537 & 0.539 & 0.641 & \cellcolor{bestblue}0.564 & \cellcolor{secondgray}0.533 \\
SLCL \cite{lin2022contrastive}                                 & -- & -- & -- & -- & -- & 0.663 & \cellcolor{bestblue}0.548 & 0.546 & 0.635 & 0.560 & 0.526 \\
ICL \cite{lin2022contrastive}                                  & -- & -- & -- & -- & -- & 0.661 & 0.542 & 0.542 & 0.634 & \cellcolor{secondgray}0.558 & 0.526 \\
SCL \cite{lin2022contrastive}                                  & -- & -- & -- & -- & -- & \cellcolor{secondgray}0.667 & \cellcolor{secondgray}0.542 & \cellcolor{bestblue}0.554 & 0.637 & \cellcolor{secondgray}0.558 & \cellcolor{bestblue}0.535 \\
\midrule
\textbf{Ours (Ambiguity)} 
& \cellcolor{bestblue}\textbf{0.163} & \cellcolor{bestblue}\textbf{0.086} & \cellcolor{bestblue}\textbf{0.837} & \cellcolor{bestblue}\textbf{0.591} & \cellcolor{bestblue}\textbf{0.622}
& \cellcolor{bestblue}\textbf{0.797} & \textbf{0.527} & \textbf{0.524}
& \cellcolor{bestblue}\textbf{0.859} & \textbf{0.508} & \textbf{0.497} \\
\bottomrule
\end{tabular}
\end{table*}

\subsection{Evaluation Metrics}
\label{subsec:metrics}

We evaluate all models using standard metrics for multi-label classification that jointly capture label-wise accuracy, ranking quality, and overall predictive performance.
Let $\mathbf{y}_i \in \{0,1\}^K$ denote the ground-truth label vector for instance $i$ and $\hat{\mathbf{y}}_i \in \{0,1\}^K$ the predicted label vector obtained by thresholding predicted probabilities $\mathbf{p}_i$.

\paragraph{Hamming Loss (HL):}
Hamming Loss measures the fraction of incorrectly predicted labels:
\begin{equation}
\mathrm{HL} = \frac{1}{NK} \sum_{i=1}^{N} \sum_{k=1}^{K} \mathbb{I}\left[\hat{y}_{ik} \neq y_{ik}\right],
\end{equation}
where lower values indicate better performance.

\paragraph{Ranking Loss (RL):}
Ranking Loss evaluates the fraction of label pairs that are incorrectly ordered:
\begin{equation}
\mathrm{RL} = \frac{1}{N} \sum_{i=1}^{N}
\frac{1}{|\mathcal{Y}_i^+||\mathcal{Y}_i^-|}
\sum_{(k,l) \in \mathcal{Y}_i^+ \times \mathcal{Y}_i^-}
\mathbb{I}\left[p_{ik} \le p_{il}\right],
\end{equation}
where $\mathcal{Y}_i^+$ and $\mathcal{Y}_i^-$ denote the sets of relevant and irrelevant labels for instance $i$, respectively.

\paragraph{Jaccard Index (Jaccard):}
The Jaccard Index measures the similarity between the predicted and ground-truth label sets for each instance:
\begin{equation}
\mathrm{Jaccard} = \frac{1}{N} \sum_{i=1}^{N}
\frac{|\hat{\mathcal{Y}}_i \cap \mathcal{Y}_i|}
     {|\hat{\mathcal{Y}}_i \cup \mathcal{Y}_i|},
\end{equation}
where $\mathcal{Y}_i = \{k \mid y_{ik}=1\}$ and $\hat{\mathcal{Y}}_i = \{k \mid \hat{y}_{ik}=1\}$ denote the sets of true and predicted labels, respectively.
The Jaccard Index directly captures set-level overlap and is particularly informative in multi-label settings where partial matches are common.
Higher values indicate better performance.

\paragraph{Micro-F1 (miF1):}
Micro-F1 aggregates true positives, false positives, and false negatives across all labels:
\begin{equation}
\mathrm{miF1} = \frac{2 \sum_k \mathrm{TP}_k}{2 \sum_k \mathrm{TP}_k + \sum_k \mathrm{FP}_k + \sum_k \mathrm{FN}_k}.
\end{equation}
This metric emphasizes performance on frequent labels and is robust under class imbalance.

\paragraph{Macro-F1 (maF1):}
Macro-F1 computes the unweighted average of per-label F1 scores:
\begin{equation}
\mathrm{maF1} = \frac{1}{K} \sum_{k=1}^{K} \mathrm{F1}_k,
\end{equation}
highlighting performance on rare emotion labels.

\paragraph{Average Precision (AP):}
Average Precision evaluates the quality of label ranking across thresholds:
\begin{equation}
\mathrm{AP} = \frac{1}{K} \sum_{k=1}^{K}
\sum_{n} (R_{k,n} - R_{k,n-1}) P_{k,n},
\end{equation}
where $P_{k,n}$ and $R_{k,n}$ denote precision and recall at threshold $n$ for label $k$.

\paragraph{Evaluation Protocol:}
All metrics are computed on fully observed test labels.
Threshold-dependent metrics (HL, miF1, maF1, Jaccard) are computed using either a fixed threshold of $0.5$ or a global threshold tuned on the development set. Ranking-based metrics (RL, AP) are computed directly from predicted probabilities without thresholding. For ablation studies, we report mean $\pm$ standard deviation over three random seeds.

\subsection{Main Results}
\textcolor{blue}{Table~\ref{tab:multilingual_results}} presents a comprehensive comparison across English, Arabic, and Spanish emotion classification benchmarks. On English, the proposed ambiguity-aware model achieves the strongest overall performance, yielding the lowest Hamming and Ranking Loss and the highest micro-F1, macro-F1, and average precision among all reported methods, indicating more accurate and better-calibrated multi-label predictions. On Arabic, our approach attains the highest micro- and macro-F1 scores, demonstrating improved robustness under partial and imbalanced supervision, while Jaccard similarity is reported only for methods that explicitly evaluate this metric. On Spanish, the proposed model achieves the highest micro-F1 and competitive macro-F1 and Jaccard similarity, despite strong performance from recent contrastive learning baselines. Overall, the results consistently suggest that explicitly modeling annotation ambiguity provides a reliable advantage across languages with heterogeneous supervision and varying degrees of label overlap.

\begin{table*}
\caption{Performance comparison of proposed models under different training settings.}
\label{tab:main_results}
\begin{tabular}{lccccc}
\toprule
Model & HL ↓ & RL ↓ & miF1 ↑ & maF1 ↑ & AP ↑ \\
\midrule
EN + $Ambiguity_{weight}$ & 0.162571 & 0.086242 & 0.837429 & 0.590508 & 0.622180 \\
ES + $Ambiguity_{weight}$ & 0.140600 & 0.111882 & 0.859400 & 0.508212 & 0.518929 \\
AR + $Ambiguity_{weight}$ & 0.202659 & 0.122450 & 0.797341 & 0.527084 & 0.538473 \\
BOTH + $Ambiguity_{weight}$ & 0.161780 & 0.100946 & 0.838220 & 0.556618 & 0.581709 \\
BOTH + $Ambiguity_{weight}$ + PU & 0.161780 & 0.100946 & 0.838220 & 0.556618 & 0.581709 \\
\bottomrule
\end{tabular}
\end{table*}
\textcolor{blue}{Table~\ref{tab:main_results}} presents the performance of the proposed framework under different training settings across individual languages and their multilingual combination. The ambiguity-weighted variant consistently yields strong results, achieving low Hamming and Ranking Loss while maintaining high micro- and macro-F1 scores, which indicates effective handling of overlapping emotion labels. Spanish attains the highest micro-F1, reflecting clearer label separability, whereas English and Arabic exhibit slightly lower but more balanced macro-F1 scores, highlighting differences in label co-occurrence and class imbalance. Training on the combined multilingual data improves robustness compared to single-language models, suggesting beneficial cross-lingual transfer. Incorporating positive-unlabeled learning does not further improve performance in this setting, indicating that ambiguity weighting already captures the dominant uncertainty present in the annotations.

\subsection{Ablation Study}

\begin{table*}
\caption{Ablation study (mean $\pm$ std over three seeds).}
\label{tab:ablation}
\begin{tabular}{lcccccc}
\toprule \midrule
Train & Mode & HL & RL & miF1 & maF1 & AP \\
\midrule \midrule
EN & Baseline & 0.1483 $\pm$ 0.0021 & 0.0876 $\pm$ 0.0005 & 0.8517 $\pm$ 0.0021 & 0.5822 $\pm$ 0.0014 & 0.6160 $\pm$ 0.0012 \\ 
EN & $Ambiguity_{weight}$ & 0.1515 $\pm$ 0.0079 & 0.0860 $\pm$ 0.0002 & 0.8485 $\pm$ 0.0079 & 0.5887 $\pm$ 0.0070 & 0.6218 $\pm$ 0.0018 \\ 
EN & Evidential & 0.2155 $\pm$ 0.0219 & 0.0963 $\pm$ 0.0004 & 0.7845 $\pm$ 0.0219 & 0.5542 $\pm$ 0.0041 & 0.5857 $\pm$ 0.0009 \\ \midrule
ES & Baseline & 0.1380 $\pm$ 0.0163 & 0.1116 $\pm$ 0.0035 & 0.8620 $\pm$ 0.0163 & 0.5081 $\pm$ 0.0106 & 0.5233 $\pm$ 0.0122 \\ 
ES & $Ambiguity_{weight}$ & 0.1332 $\pm$ 0.0053 & 0.1068 $\pm$ 0.0038 & 0.8668 $\pm$ 0.0053 & 0.5247 $\pm$ 0.0119 & 0.5393 $\pm$ 0.0144 \\ 
ES & Evidential & 0.3119 $\pm$ 0.0212 & 0.1588 $\pm$ 0.0034 & 0.6881 $\pm$ 0.0212 & 0.4037 $\pm$ 0.0077 & 0.3804 $\pm$ 0.0067 \\ \midrule
AR & Baseline & 0.2345 $\pm$ 0.0411 & 0.1210 $\pm$ 0.0081 & 0.7655 $\pm$ 0.0411 & 0.5268 $\pm$ 0.0102 & 0.5425 $\pm$ 0.0168 \\ 
AR & $Ambiguity_{weight}$ & 0.2000 $\pm$ 0.0039 & 0.1165 $\pm$ 0.0046 & 0.8000 $\pm$ 0.0039 & 0.5362 $\pm$ 0.0070 & 0.5556 $\pm$ 0.0126 \\ 
AR & Evidential & 0.3385 $\pm$ 0.0361 & 0.1377 $\pm$ 0.0048 & 0.6615 $\pm$ 0.0361 & 0.5022 $\pm$ 0.0055 & 0.4918 $\pm$ 0.0096 \\ \midrule
BOTH & Baseline & 0.1411 $\pm$ 0.0023 & 0.0907 $\pm$ 0.0004 & 0.8589 $\pm$ 0.0023 & 0.5890 $\pm$ 0.0014 & 0.6191 $\pm$ 0.0022 \\ 
BOTH & $Ambiguity_{weight}$ & 0.1412 $\pm$ 0.0046 & 0.0897 $\pm$ 0.0008 & 0.8588 $\pm$ 0.0046 & 0.5909 $\pm$ 0.0006 & 0.6229 $\pm$ 0.0007 \\ 
BOTH & Evidential & 0.1579 $\pm$ 0.0089 & 0.1022 $\pm$ 0.0010 & 0.8421 $\pm$ 0.0089 & 0.5575 $\pm$ 0.0026 & 0.5802 $\pm$ 0.0013 \\ \midrule
\bottomrule
\end{tabular}
\end{table*}
\textcolor{blue}{Table~\ref{tab:ablation}} reports an ablation study across languages, averaged over three random seeds. Compared to the baseline, the ambiguity-weighted training consistently improves macro-F1 and average precision across all languages, indicating better handling of label overlap and class imbalance, while maintaining competitive Hamming and Ranking Loss. The gains are most pronounced for Spanish and Arabic, where annotation sparsity and imbalance are more severe. In contrast, the evidential variant shows substantially degraded performance, with higher error rates and lower F1 scores, suggesting that uncertainty modeling alone is insufficient without explicitly accounting for missing and ambiguous labels. When training on the combined multilingual data, ambiguity weighting again provides the most stable trade-off between accuracy and robustness, confirming its effectiveness across both monolingual and multilingual settings.
\subsection{Further Analysis}

\begin{table*}
\caption{Performance stratified by ambiguity level (entropy $H$) on Spanish validation/test instances. $w=\exp(-\tau H)$; higher $H$ indicates more ambiguous instances.}
\label{tab:ambig_perf_es}
\begin{tabular}{c l r c c c c}
\toprule
Bin & $H$ range & $\#Inst$ & $H$ mean & $w$ mean & miF1 $\uparrow$ & AP $\uparrow$ \\
\midrule
1 & [0.1059, 0.1628] & 136 & 0.1443 & 0.7496 & 0.8424 & 0.6118 \\
2 & [0.1628, 0.1907] & 136 & 0.1770 & 0.7019 & 0.6903 & 0.6117 \\
3 & [0.1907, 0.2207] & 135 & 0.2059 & 0.6626 & 0.6284 & 0.6019 \\
4 & [0.2207, 0.2600] & 136 & 0.2405 & 0.6184 & 0.5830 & 0.5652 \\
5 & [0.2600, 0.3706] & 136 & 0.2897 & 0.5607 & 0.4948 & 0.4594 \\
\bottomrule
\end{tabular}
\end{table*}
\textcolor{blue}{Table~\ref{tab:ambig_perf_es}} analyzes model performance on Spanish instances stratified by ambiguity level, measured via label entropy. As ambiguity increases, both micro-F1 and average precision consistently decline, reflecting the greater difficulty of instances with more uncertain or overlapping emotion annotations. At the same time, the mean instance weight decreases monotonically with higher entropy, confirming that the ambiguity-weighted scheme effectively down-weights highly ambiguous examples during training. Notably, performance degradation is gradual rather than abrupt, suggesting that the model remains robust across a wide range of ambiguity levels while prioritizing more reliable supervision. This behavior supports the intended role of entropy-based weighting in mitigating the impact of annotation uncertainty without discarding ambiguous instances entirely.

\begin{table}
\caption{Label-wise uncertainty on Arabic split. Higher entropy indicates more ambiguous emotion labels.}
\label{tab:label_uncertainty_ar}
\begin{tabular}{l c c c c}
\toprule
Emotion & $\mathbb{E}[H_k]$ & $\mathrm{Std}(H_k)$ & $\mathbb{E}[p_k]$ & $\mathrm{Std}(p_k)$ \\
\midrule
pessimism & 0.3596 & 0.2441 & 0.1871 & 0.1912 \\
sadness & 0.2950 & 0.2192 & 0.3751 & 0.3803 \\
anticipation & 0.2314 & 0.1453 & 0.0860 & 0.1264 \\
anger & 0.2314 & 0.2182 & 0.3227 & 0.3870 \\
love & 0.2293 & 0.2137 & 0.3223 & 0.3882 \\
disgust & 0.2289 & 0.2190 & 0.1425 & 0.2324 \\
optimism & 0.2285 & 0.2305 & 0.3112 & 0.3809 \\
fear & 0.1794 & 0.1717 & 0.1769 & 0.3110 \\
joy & 0.1734 & 0.1973 & 0.3358 & 0.4163 \\
trust & 0.1694 & 0.1650 & 0.0575 & 0.0925 \\
surprise & 0.0929 & 0.0904 & 0.0230 & 0.0424 \\
\bottomrule
\end{tabular}
\end{table}
\textcolor{blue}{Table~\ref{tab:label_uncertainty_ar}} presents label-wise uncertainty statistics for the Arabic split, measured via the expected entropy of each emotion. Emotions such as pessimism and sadness exhibit the highest average entropy and variance, indicating substantial annotation ambiguity and inconsistent label assignment across instances. In contrast, surprise and trust show markedly lower entropy values, suggesting more consistent labeling and clearer semantic cues. The expected label probabilities further reveal that highly ambiguous emotions also tend to have greater variability in prevalence, reflecting both semantic overlap and annotation sparsity. These findings underscore the heterogeneous uncertainty structure across emotion categories and motivate label-adaptive weighting to prevent highly ambiguous emotions from disproportionately influencing model updates.

\begin{table}
\caption{\small Training stability across seeds (mean $\pm$ std) for different uncertainty modes.}
\label{tab:stability}
\resizebox{\columnwidth}{!}{
\begin{tabular}{l l r c c} 
\toprule
Train & Mode & $n$ & miF1 $\uparrow$ & AP $\uparrow$ \\
\midrule
EN & $ambiguity_{weight}$ & 3 & 0.8485 $\pm$ 0.0096 & 0.6218 $\pm$ 0.0022 \\
EN & baseline & 3 & 0.8517 $\pm$ 0.0026 & 0.6160 $\pm$ 0.0014 \\
EN & evidential & 3 & 0.7845 $\pm$ 0.0268 & 0.5857 $\pm$ 0.0011 \\
ES & $ambiguity_{weight}$ & 3 & 0.8668 $\pm$ 0.0065 & 0.5393 $\pm$ 0.0177 \\
ES & baseline & 3 & 0.8620 $\pm$ 0.0199 & 0.5233 $\pm$ 0.0150 \\
ES & evidential & 3 & 0.6881 $\pm$ 0.0260 & 0.3804 $\pm$ 0.0082 \\
AR & $ambiguity_{weight}$ & 3 & 0.8000 $\pm$ 0.0048 & 0.5556 $\pm$ 0.0154 \\
AR & baseline & 3 & 0.7655 $\pm$ 0.0504 & 0.5425 $\pm$ 0.0205 \\
AR & evidential & 3 & 0.6615 $\pm$ 0.0442 & 0.4918 $\pm$ 0.0118 \\
BOTH & $ambiguity_{weight}$ & 3 & 0.8588 $\pm$ 0.0056 & 0.6229 $\pm$ 0.0009 \\
BOTH & baseline & 3 & 0.8589 $\pm$ 0.0028 & 0.6191 $\pm$ 0.0027 \\
BOTH & evidential & 3 & 0.8421 $\pm$ 0.0109 & 0.5802 $\pm$ 0.0016 \\
\bottomrule
\end{tabular}}
\end{table}
\textcolor{blue}{Table~\ref{tab:stability}} reports training stability across three random seeds for different uncertainty modeling strategies. The ambiguity-weighted approach consistently achieves strong performance with low variance across all languages, indicating stable optimization and reliable convergence. Compared to the baseline, ambiguity weighting yields comparable or improved mean micro-F1 and average precision, while often reducing performance variability, particularly for Spanish and Arabic. In contrast, the evidential setting exhibits substantially higher variance and lower mean performance, suggesting sensitivity to initialization and difficulty in calibrating uncertainty under partial supervision. Similar trends are observed in the multilingual setting, where ambiguity weighting provides a favorable balance between accuracy and stability, reinforcing its suitability for multilingual emotion classification with heterogeneous annotations.

\subsection{Interpretability}

\begin{figure*}[t]
    \centering

    \begin{subfigure}[t]{\columnwidth}
        \centering
        \includegraphics[width=\columnwidth]{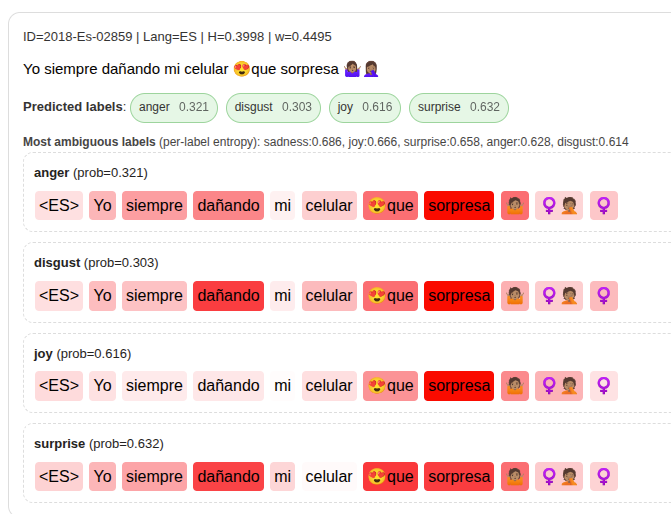}
        \caption{\textbf{Spanish} (high ambiguity; emoji-driven cues).}
        \label{fig:interp_es}
    \end{subfigure}

    \begin{subfigure}[t]{\columnwidth}
        \centering
        \includegraphics[width=\columnwidth]{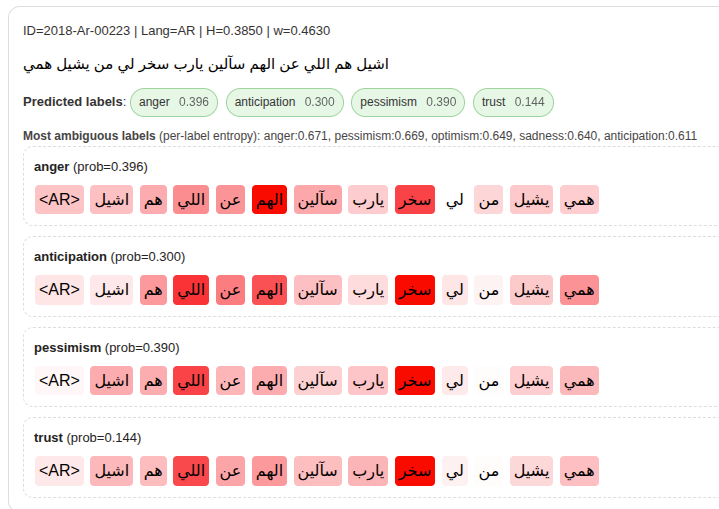}
        \caption{\textbf{Arabic} (context-driven ambiguity; morphology-sensitive cues).}
        \label{fig:interp_ar}
    \end{subfigure}
    \begin{subfigure}[t]{\columnwidth}
        \centering
        \includegraphics[width=\columnwidth]{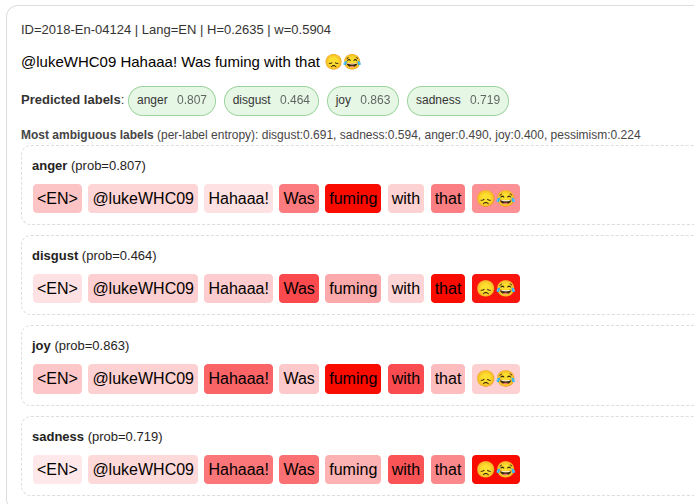}
        \caption{\textbf{English} (multi-label co-occurrence with clear lexical cues).}
        \label{fig:interp_en}
    \end{subfigure}

    \caption{
    \textbf{Interpretability under ambiguity (test set).}
    For each instance, we report predicted labels with probabilities, the ambiguity score (entropy $H$), and the derived training influence weight $w=\exp(-\tau H)$.
    Token-level highlights are computed using Gradient~$\times$~Input for each predicted label.
    Darker red indicates higher token contribution to the corresponding emotion.
    Across languages, ambiguous instances yield higher entropy and lower weights, while token attributions identify salient lexical, emoji, or contextual cues supporting multi-label predictions.
    }
    \label{fig:interp_examples}
\end{figure*}

\textcolor{blue}{Figure~\ref{fig:interp_examples}} presents qualitative interpretability results on held-out test instances across Spanish, Arabic, and English.
Each example reports multi-label predictions, the instance-level ambiguity score $H$, and the corresponding influence weight $w=\exp(-\tau H)$, which reflects how strongly an instance would contribute during ambiguity-weighted learning.
Token-level explanations are obtained using Gradient~$\times$~Input, producing label-specific attribution heatmaps.

In Spanish (\textcolor{blue}{Figure~\ref{fig:interp_es}}), emojis and mixed affect lead to multiple plausible emotions (e.g., joy, surprise, anger, disgust), yielding higher entropy and reduced weight.
In Arabic (\textcolor{blue}{Figure~\ref{fig:interp_ar}}), ambiguity arises primarily from contextual and morphological cues rather than explicit affective markers, resulting in competing emotions (e.g., anger and pessimism) and moderate uncertainty.
In English (\textcolor{blue}{Figure~\ref{fig:interp_en}}), explicit lexical signals and emoticons support multiple co-occurring emotions with comparatively lower entropy and higher weight.
Overall, these examples demonstrate that the proposed framework provides transparent multi-label reasoning, quantifies ambiguity explicitly, and produces coherent token-level evidence across languages.

\subsection{Case Study}
\begin{table*}[t]
\centering
\small
\caption{ \small Nearest-neighbor explanations using cosine similarity in the learned embedding space. For each development-set query, we report the most similar training instance along with the corresponding emotion labels. 
Texts are lightly paraphrased for clarity and anonymization.
}
\setlength{\tabcolsep}{5pt}
\renewcommand{\arraystretch}{1.25}
\resizebox{\textwidth}{!}{
\begin{tabular}{l p{5.20cm} p{5.20cm} c p{4.2cm}}
\toprule
\textbf{Lang} & \textbf{Dev Query (paraphrased)} & \textbf{Nearest Train Instance (paraphrased)} & \textbf{Sim.} & \textbf{Train Emotion Labels} \\
\midrule
EN & 5,000 new US \#citizens swearing in now! \#tears \#congratulations & Thank you @twitter for the balloons today.  \#goodday \#48 & 1.00 & joy,optimism \\
EN & 'What has you smiling like that' 'Dominos sent me a coupon' & So is texting a guy 'I'm ready for sex now' considered flirting?' \#shocking & 1.00 & anger,anticipation,disgust,surprise \\
EN & @ehtesham\_toi @RaviShastriOfc unfortunate decision, our players are more of superstars than cricketers. BCCI scared of their tantrums. & @cotsonika @NHL well, they should all be unhappy for the way they played.  Right @fmjacob ?? & 1.00 & anger,disgust,sadness \\
EN & @Rainbow6Game felt guilt from playing this game. \#horrible \#RainbowSixSiege & @Yoshi\_OnoChin can you please not have Canadian players play US players, that lag  is atrocious. \#fixthisgame \#trash \#sfvrefund  \#rage & 1.00 & anger,disgust \\
EN & @BadHombreNPS @SecretaryPerry If this didn't make me so angry, I'd be laughing at this tweet! & @TrueAggieFan oh so that's where Brian was! Where was my invite? \#offended & 1.00 & anger,disgust \\
EN & today i feel like an exposed wire dangerously close to another exposed wire and any provocation will fry me & what's the nicest way to tell someone cheerfully whistling outside my apartment door that I will end them should they continue to whistle & 1.00 & anger,disgust \\
EN & Are u \#depressed \#hypo \#manic \#lonely \#bored \#nofriends \#needfreinds \#friend I feel chatty I wanna help ppl or just \#makefriend 's \#dm \#moms & Bored rn leave Kik/Snapchat \#kik \#kikme \#kikmessage \#boredaf \#bored \#snapchatme \#snapchat \#snap \#country \#countrygirl & 1.00 & disgust \\
EN & +++ '\#Dearly \#beloved, avenge not yourselves, but rather give place unto \#wrath: for it is \#written, \#Vengeance is \#mine; I …' \#Romans12v19 & Do not despise the Lord ’s \#discipline, do not \#resent his rebuke, because the \#Lord disciplines those he loves... \#Proverbs 3:11-12 & 1.00 & optimism \\
\bottomrule
\end{tabular}}
\label{tab:nn_examples}
\end{table*}
\textcolor{blue}{Table~\ref{tab:nn_examples}} presents qualitative nearest-neighbor explanations based on cosine similarity in the learned embedding space. For each development instance, the model retrieves a highly similar training example that shares both semantic content and emotion structure, providing insight into how predictions are grounded in prior observations. The retrieved neighbors consistently exhibit overlapping or closely related emotion labels, such as anger and disgust or joy and optimism, illustrating the model’s ability to capture nuanced co-occurrence patterns rather than relying on single-label cues. Importantly, even for emotionally complex or metaphorical expressions, the nearest neighbors reflect comparable affective intent, suggesting that the learned representations align well with human-interpretable emotional similarity. This case study highlights how ambiguity-aware training encourages coherent embedding geometry, enabling transparent instance-level explanations through similarity-based retrieval.

\section{Discussion}\label{discussion}
The experimental results consistently demonstrate that explicitly modeling annotation ambiguity is critical for multi-label emotion classification under partial supervision. Across languages and training settings, ambiguity-weighted learning yields a favorable balance between accuracy and robustness, improving macro-level performance and average precision while maintaining low error rates. These gains are particularly evident in languages and labels with higher uncertainty, where treating unannotated labels as negative evidence can otherwise bias learning.

The ablation and stability analyses further clarify the role of uncertainty modeling. While evidential approaches aim to capture predictive uncertainty, they struggle in the presence of systematically missing labels, leading to unstable training and degraded performance. In contrast, entropy-based ambiguity weighting directly addresses the structure of the supervision signal by down-weighting instances with high label uncertainty, resulting in more stable optimization across random seeds and datasets. This distinction highlights the importance of aligning uncertainty modeling with the annotation process rather than relying solely on predictive variance. Qualitative analyses provide additional insight into model behavior. The nearest-neighbor case studies show that ambiguity-aware training induces a coherent embedding space in which emotionally similar instances cluster together, enabling interpretable, instance-level explanations. Moreover, label-wise uncertainty analysis reveals that different emotions exhibit markedly different ambiguity profiles, suggesting that uniform treatment of labels is suboptimal and that adaptive strategies are necessary.

Overall, these findings suggest that ambiguity is not merely noise to be suppressed but a structural property of emotion annotations that should be explicitly modeled. By incorporating uncertainty-aware weighting into the learning objective, the proposed framework better reflects the realities of multilingual emotion annotation and offers a principled path toward more reliable and interpretable multi-label classifiers.

\section{Limitations and Future Works}\label{Limitations_futurework}
Despite its usefulness, the suggested framework has a number of drawbacks. First, ambiguity is represented by entropy derived from observable label distributions, which is dependent on annotation quality and may miss deeper semantic uncertainty or annotator disagreement patterns. Second, the method has only been validated for emotion classification benchmarks using brief, user-generated texts; its applicability to lengthier papers or other multi-label tasks has not yet been explored, despite the fact that it generalizes well across the languages under investigation. Furthermore, similarity-based explanations may be less helpful in situations with limited resources or extreme imbalance since they rely on the availability of comparable training instances.  To enable researchers to provide more expressive and accurate explanations under ambiguity, future work will explore integration with large language models, extend the framework to other multilingual and multimodal classification tasks, and investigate richer uncertainty modeling that incorporates annotator-level information.

\section{Conclusions}\label{conclusion}
The article presents an ambiguity-aware framework for multilingual multi-label emotion classification under partial supervision. The proposed method reduces the impact of missing and overlapping emotion labels by explicitly quantifying annotation uncertainty with entropy-based instance weighting, rather than assuming unobserved labels are negative. Extensive research in English, Spanish, and Arabic demonstrates continuous improvements in performance, stability, and robustness compared to rival baselines and other uncertainty modeling approaches. Further research demonstrates that ambiguity weighting naturally adapts to different levels of label uncertainty, resulting in more coherent representation spaces that facilitate interpretable, instance-level explanations. Overall, the findings emphasize the necessity of matching learning objectives to the structure of real-world annotations, and they imply that ambiguity-aware training provides a practical and successful foundation for trustworthy multilingual emotion analysis.

\section*{Funding}
No external funding was received.

\section*{Acknowledgments}
We would like to thank the Bangladesh University of Business and Technology (BUBT) for providing the necessary experimental tools and support that facilitated this research.

\section*{Conflicts of Interest}
The authors declare that they have no conflict of interest.

\section*{Data Availability Statement}
The dataset we used in this study is SemEval-2018 Task 1 Emotion Classification (E-C) and the implementations code is available at \url{https://github.com/MIthun667/Reasoning-under-Ambiguity}

\printcredits

\bibliographystyle{model1-num-names}
\bibliography{cas-refs}
\end{document}